\documentclass[10pt, a4paper]{article}
\usepackage{lrec2022} 
\usepackage{multibib}
\newcites{languageresource}{Language Resources}
\usepackage{graphicx}
\usepackage{tabularx}
\usepackage{soul}
\usepackage{titlesec}
\titleformat{\section}{\normalfont\large\bfseries\center}{\thesection.}{1em}{}
\titleformat{\subsection}{\normalfont\SmallTitleFont\bfseries\raggedright}{\thesubsection.}{1em}{}
\titleformat{\subsubsection}{\normalfont\normalsize\bfseries\raggedright}{\thesubsubsection.}{1em}{}
\renewcommand\thesection{\arabic{section}}
\renewcommand\thesubsection{\thesection.\arabic{subsection}}
\renewcommand\thesubsubsection{\thesubsection.\arabic{subsubsection}}
\newcommand\CORPUSNAME{SDS-200}
\usepackage{epstopdf}
\usepackage[utf8]{inputenc}

\usepackage{hyperref}
\usepackage{xstring}

\usepackage{color}

\usepackage{multirow,tabularx}
\usepackage{booktabs}

\title{\CORPUSNAME: A Swiss German Speech to Standard German Text Corpus}

\name{{\bf \large Michel Plüss}\textsuperscript{a}, {\bf \large Manuela Hürlimann}\textsuperscript{b}, {\bf \large Marc Cuny}\textsuperscript{c}, {\bf \large Alla Stöckli}\textsuperscript{c}, {\bf \large Nikolaos Kapotis}\textsuperscript{b}, \\
{\bf \large Julia Hartmann}\textsuperscript{a}, {\bf \large Malgorzata Anna Ulasik}\textsuperscript{b}, {\bf \large Christian Scheller}\textsuperscript{a}, {\bf \large Yanick Schraner}\textsuperscript{a},  \\
{\bf \large Amit Jain}, {\bf \large Jan Deriu}\textsuperscript{b}, {\bf \large Mark Cieliebak}\textsuperscript{b}\textsuperscript{c}, {\bf \large Manfred Vogel}\textsuperscript{a}} 
\address{
\textsuperscript{a}University of Applied Sciences and Arts Northwestern Switzerland, Windisch  \\
\textsuperscript{b}Zurich University of Applied Sciences, Winterthur \\
\textsuperscript{c}SpinningBytes AG, Winterthur \\michel.pluess@fhnw.ch}

\abstract{
We present \CORPUSNAME, a corpus of Swiss German dialectal speech with Standard German text translations, annotated with dialect, age, and gender information of the speakers. The dataset allows for training speech translation, dialect recognition, and speech synthesis systems, among others. The data was collected using a web recording tool that is open to the public. Each participant was given a text in Standard German and asked to translate it to their Swiss German dialect before recording it. To increase the corpus quality, recordings were validated by other participants. The data consists of 200 hours of speech by around 4000 different speakers and covers a large part of the Swiss German dialect landscape. We release \CORPUSNAME~alongside a baseline speech translation model, which achieves a word error rate (WER) of 30.3 and a BLEU score of 53.1 on the \CORPUSNAME~test set. Furthermore, we use \CORPUSNAME~to fine-tune a pre-trained XLS-R model, achieving 21.6 WER and 64.0 BLEU.
 \\ \newline \Keywords{Corpus, Less-Resourced/Endangered Languages, Speech Recognition/Understanding, Speech Resource/Database, Statistical and Machine Learning Methods} }

\begin{document}

\maketitleabstract

\section{Introduction}
We present Schweizer Dialektsammlung~(\CORPUSNAME), a corpus of Swiss German dialectal speech with the corresponding Standard German text. The data consists of 200 hours of speech. We make the corpus publicly available~\footnote{\url{https://swissnlp.org/datasets/}}.

Swiss German is a family of German dialects spoken by around five million people in Switzerland. It differs from Standard German regarding phonetics, vocabulary, morphology, and syntax and is primarily a spoken language. While it is also used in writing, particularly in informal text messages, it lacks a standardized orthography. This leads to difficulties for automated text processing due to spelling ambiguities and huge vocabulary size. Therefore, it is often preferable to work with Standard German text, for which automated processing tools exist in abundance. The main challenge is that Swiss German is not a unified language but a collection of dialects, which sometimes differ significantly in phonetics, grammar, and vocabulary. The immense vocabulary makes it hard to create a Swiss German Automatic Speech Recognition (ASR) system. Due to these reasons, Swiss German is a low-resource language. One way to tackle Swiss German ASR is an end-to-end Swiss German speech to Standard German text approach. This can be viewed as a speech translation (ST) task with similar source and target languages.

Training a model for this task requires a substantial amount of data. Unfortunately, not enough public data is available for Swiss German. The largest available corpus, the Swiss Parliaments Corpus (SPC) \cite{pluess2021a}, is limited to the Bernese dialect. However, there are many different dialects in Switzerland, some of which differ substantially from Bernese because the difference between dialects can be significant, especially regarding vocabulary and pronunciation; as many dialects as possible should be part of the training data.

For \CORPUSNAME, we created a web recording tool\footnote{\url{https://dialektsammlung.ch/de}} which is open to the public. The idea is that the public can record Standard German sentences in their Swiss German dialect. Other participants then validate the recordings. Almost 4000 different participants from all over Switzerland helped create a high-quality corpus covering a large part of the Swiss German dialect landscape. To cover a wide range of topics and increase vocabulary diversity, we used texts from Swiss newspapers and the German Common Voice corpus. The code of the tool is open source\footnote{\url{https://github.com/stt4sg/dialektsammlung-public}}.

The remainder of this paper is structured as follows: Related work is discussed in section 2. The data collection process is described in section 3. Corpus preparation and statistics can be found in section 4. In section 5, we describe a baseline model trained on the corpus. Section 6 wraps up the paper and gives directions for future work.

\section{Related Work}

End-to-end approaches are widely used in deep learning, especially natural language processing (NLP).
In the domain of speech translation, suitable corpora are scarce.
The MuST-C dataset \cite{di-gangi-etal-2019-must} provides ~400 h of English speech data with sentence-aligned text for eight different languages (German, French, Spanish, Italian, Dutch, Portuguese, Romanian, and Russian).
The MuST-C data is collected from TED talks, providing a variety of topics and speakers (male/female, native/non-native speakers).
TED talks are manually transcribed and translated, providing a high-quality data source.

Europarl \cite{9054626} is another ST corpus with speech and sentence-aligned text for 6 European languages (English, German, French, Spanish, Italian, and Portuguese) containing between 20 and 89 hours of audio for 30 pairs.
The sentence alignment is done automatically.
Due to the automatic alignment, audio data with low alignment confidence is discarded, and the data quality is lower than manual text alignment.
Europarl contains speeches held in the European Parliament.

Four public datasets contain Swiss German audio with transcripts. 
SPC \cite{pluess2021a} is the largest corpus with 293 hours of data in the Bernese dialect recorded in the Bernese cantonal parliament.
The text and audio are automatically aligned by using commercial Standard German ASR systems, followed by a forced sentence alignment using the Needleman-Wunsch algorithm \cite{Needleman-Wunsch}.
The ArchiMob dataset \cite{Scherrer2019} includes 69 hours of Swiss German speech and Swiss German transcript.
There are no Standard German transcripts available.
The Radio Rottu Oberwallis dataset  \cite{Garner2014b} includes 8 hours of speech, 2 of which are provided with Standard German transcripts.
SwissDial \cite{Dogan2021} is a high-quality dataset including eight different Swiss German dialects with roughly 3 hours of audio data per dialect.
The sentences are crawled from newspapers and Wikipedia and then manually translated into the selected eight Swiss German dialects.
The translated sentences are then recorded sentence by sentence in a studio setting.

\CORPUSNAME~combines the strengths of the existing corpora in Swiss German ASR with a large size of 200 hours, Standard German transcripts, and perfect alignment. What makes it unique is the coverage of a large part of the Swiss German dialect landscape and that almost 4000 different speakers made the recordings. We now describe the components in more detail.

\section{Data Collection}

\begin{figure*}[!t]
\begin{center}
\includegraphics[width=1.0\textwidth]{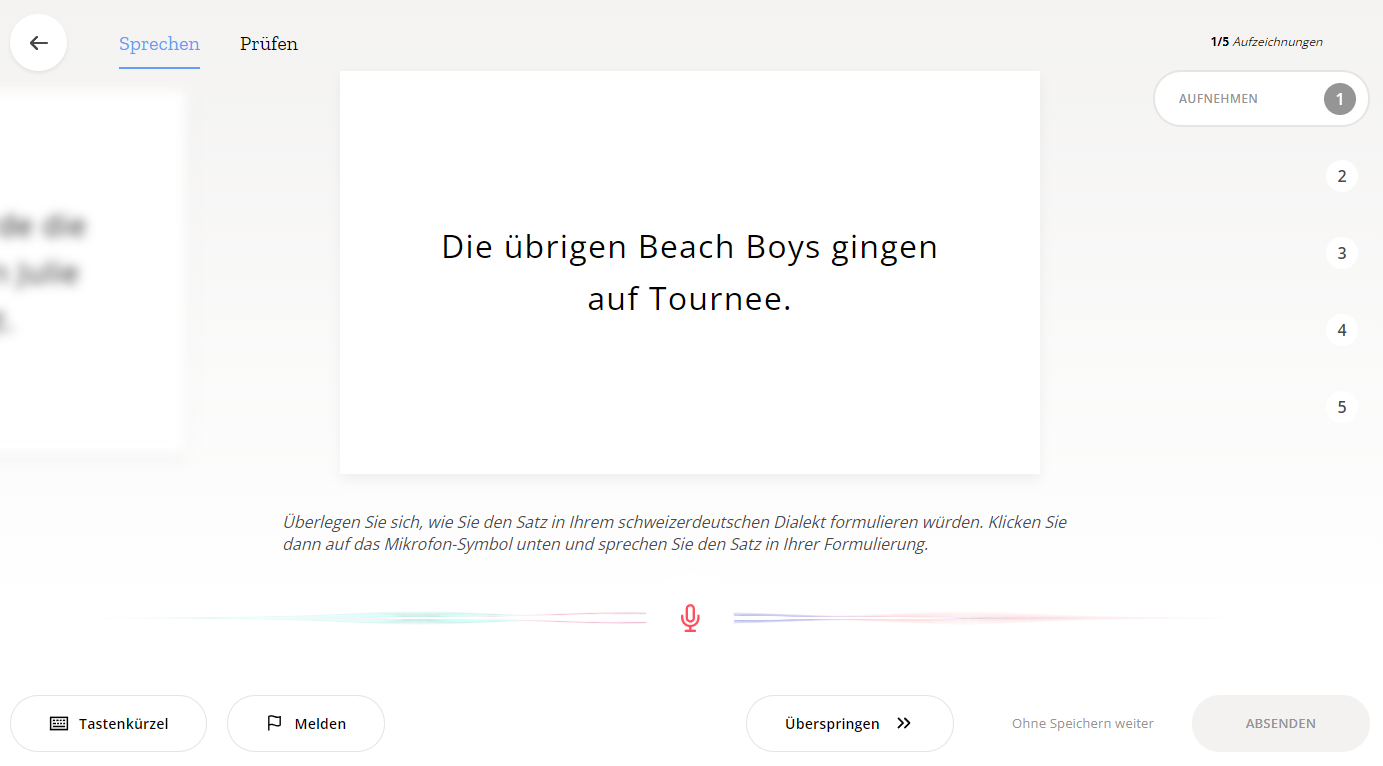}
\caption{Recording step in our tool. "\emph{Die übrigen Beach Boys gingen auf Tournee.}" is the sentence to be recorded.}
\label{fig.1}
\end{center}
\end{figure*}

\begin{figure*}[!t]
\begin{center}
\includegraphics[width=1.0\textwidth]{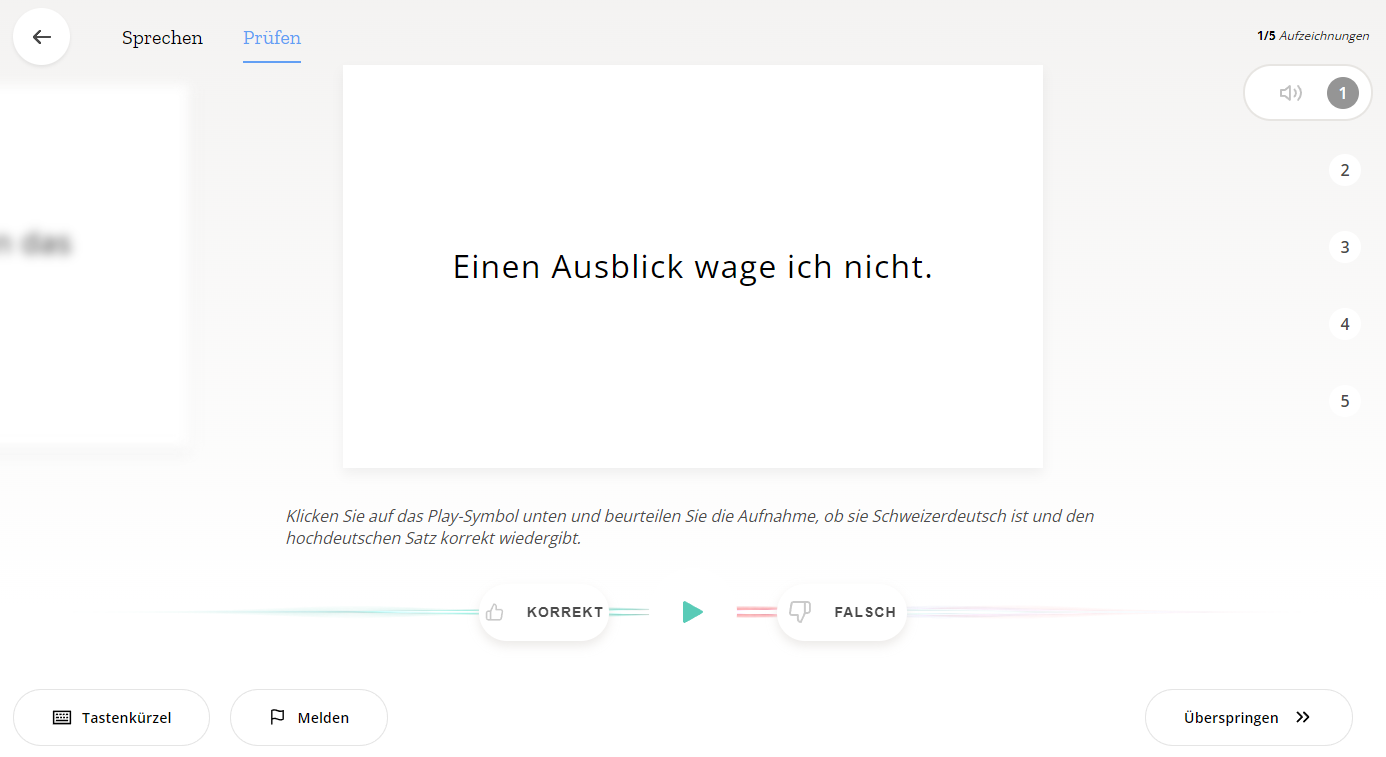}
\caption{Validation step in our tool. "\emph{Einen Ausblick wage ich nicht.}" is the Standard German sentence. The recording must be played and then judged as correct ("\emph{Korrekt}") or wrong / inaccurate ("\emph{Falsch}").}
\label{fig.2}
\end{center}
\end{figure*}

Our data collection tool is based on the Common Voice platform~\cite{ardila2020}. We adapted the annotation guidelines to the special case of Swiss German. We use the two-step annotation process of the original platform consisting of a recording step and a validation step (see Figures \ref{fig.1} and \ref{fig.2}). For the recording step, we presented Standard German sentences from Swiss newspapers, covering diverse topics and Switzerland-specific named entities, and texts from the German Common Voice corpus to the participants. They were then asked to translate each sentence into their Swiss German dialect and record it. For the validation step, the participants were presented with a sentence-recording pair and asked if the recording contained an accurate Swiss German translation of the Standard German sentence.

The goal was to create a corpus with as many hours and as much dialect and topic diversity as possible. We worked extensively with the Swiss media to reach as many people as possible. To enhance the engagement, we organized two contests on our platform.  The leaderboard contest awarded prices to the participants with the most recordings, factoring in the quality of their translations. The \emph{Clash of Cantons} contest was a competition between the 26 Swiss cantons.

\subsection{Sentence Selection}
The sentences used for the recordings were derived from Swiss newspapers and the German dataset of Common Voice. We used newspaper articles from all categories from the past five years. As the speakers' task consisted of translating the sentences from Standard German to Swiss German, not just reading them, we expected the speakers' cognitive effort to be larger, hence the error probability to be higher. Keeping this in mind, we carefully selected sentences to ensure lexical diversity and reduce sentence complexity. To this end, we selected only sentences between 5 and 12 tokens long. We applied the following filtering criteria:

\begin{itemize}
\setlength\itemsep{0em}
    \item Exclude sentences containing tokens that occur less than 1000 times per billion words. We use the Exquisite Corpus\footnote{\url{https://github.com/LuminosoInsight/exquisite-corpus}} to compute the word frequencies.
    \item Exclude sentences with a large number of rare words having an average word frequency below 10'000 per billion words.
    \item We removed sentences with dates and numbers with more than three digits. This is to reduce inconsistencies in how speakers read or translate the prompts.
    \item Sentences containing citations, e-mail addresses, hashtags, and phrases in brackets are also removed. 
    \item We kept only complete sentences. We used simple heuristics to remove incomplete sentences. For instance, each sentence begins with an uppercase letter or a digit, and a sentence should contain at least one noun, pronoun, or proper noun and one verb.  
\end{itemize}

The final set of prompts contains 1'267'195 sentences. Our tool samples newspaper sentences in 80\% of cases, and in 20\% of cases, it samples from the German Common Voice pool. 

\subsection{Recording Tool}

We made two adaptions to the original Common Voice \cite{ardila2020} platform. First, we added the possibility for the participants to specify the zip code of origin of their dialect\footnote{The origin of a participant's dialect could for example be the place where he or she grew up and / or went to school. The specified zip code is not to be confused with the current place of residence, which would not allow reliable inference of a participant's dialect.}. This allows us to investigate dialects in different granularity levels: coarse dialect regions, cantons, fine-grained dialect regions, and even individual municipalities. Additional demographic information such as age and gender selection is already available in Common Voice. Second, we adapt the annotation guidelines to cover the special case of Swiss German. The annotation is performed in two steps: a recording step and a validation step.

\noindent{\textbf{Step 1: Recording.}} During the recording step, depicted in Figure \ref{fig.1}, the participant is shown a Standard German sentence and asked to translate it to Swiss German speech. Sentences are recorded in packages of 5 and can be skipped or reported if necessary. One crucial point for our Swiss German speech to Standard German text use case is the inherent translation step the participant has to do before recording. As an example, the participant is presented with the following Standard German sentence: "\emph{Robben verstand dies wie viele andere Spieler nicht.}". The participant should then think about how he or she would formulate this sentence in his or her Swiss German dialect, e.g. "\emph{De Robben het das wie vieli anderi Spieler nid verstande.}", before actually recording the Swiss German version. This can include vocabulary as well as grammar changes, such as changing the past tense from Standard German "\emph{verstand}" to Swiss German "\emph{het (...) verstande}", which is necessary because the imperfect tense does not exist in Swiss German, where the perfect tense is used instead. We display an explanation popup with examples before the first recording to make this clear to participants. We also display a short explanation below the sentence to be recorded (see Figure \ref{fig.1}).

\noindent{\textbf{Step 2: Validation.}} Figure \ref{fig.2} depicts the validation function. Participants are asked to listen to other  recordings and judge whether the recording contains an accurate Swiss German translation of the Standard German sentence. Recordings are again validated in packages of 5 and can be reported or skipped if necessary. Similar to the recording function, we display a detailed explanation with examples of wrong (e.g. recording is in Standard German rather than Swiss German) or inaccurate (e.g. wrong tense) translations when a participant visits the validation page for the first time.

\subsection{Collection Process}

\begin{table}[t]
    \centering
    \begin{tabular}{lrrr}
        \toprule
        \textbf{Split} & \textbf{Hours} & \textbf{Sentences} & \textbf{Speakers} \\
        \midrule
        train (raw) & $188.9$ & $144'468$ & $3428$ \\
        train (filtered) & $178.3$ & $135'271$ & $3247$ \\
        validation & $5.2$ & $3638$ & $288$ \\
        test & $5.4$ & $3636$ & $281$ \\
        \bottomrule
    \end{tabular}
    \caption{Data splits of the Dialektsammlung corpus.}
    \label{tab:splits}
\end{table}

To reach as many people as possible, we collaborated with a range of national and local newspapers, television networks, and radio stations. In addition, four well-known Swiss comedians agreed to record a short video supporting the project and share it on their social media accounts, some of them reaching more than 100'000 followers.

To keep the participants motivated, we organized two contests, the leaderboard contest and the \emph{Clash of Cantons}.

\noindent{\textbf{Leaderboard.}} The leaderboard contest was a competition between all registered participants. For each participant, we computed a score based on the number of recordings, the number of validations given, and the number of positive validations received. The top ten of the leaderboard were awarded attractive Switzerland-themed prizes. Furthermore, the participant with the highest recording quality (lowest rejection rate) was awarded a special prize. 

\noindent{\textbf{Clash of Cantons.}} The \emph{Clash of Cantons} was a competition between the 26 Swiss cantons. The idea was to spark a competition between the cantons and for participants to "fight" for their respective canton. The winning canton was picked according to its number of recordings, weighted by their average quality, normalized by the population of the canton.

The data of the corpus described here was collected over seven months, with 58 \% of recordings made during the 38 days where the two contests were held. The current version contains 200 hours of raw speech data in MP3 format with a sampling rate of 32 kHz.

\section{Corpus Preparation and Data Statistics}

\subsection{Data filtering}

Crowd-sourced data needs filtering to ensure high data quality. 
We used the public validation process to filter bad samples such as empty, truncated, or silent recordings and wrong translations.

Of all recorded data, 33\% have been validated, and of these samples, 88\% have been accepted.
To also use a large amount of unvalidated samples, we allow unvalidated samples as well under the following conditions:

\begin{itemize}
    \item The speaker \emph{has some} validated recordings and more than 80\% of the validated clips are accepted. 
    \item The speaker \emph{has no} validated recordings and the duration is within 2 to 12 seconds. 
\end{itemize}

We found that we were able to filter out many clips with recording problems (e.g., empty recordings) with the second rule.
Since the added unvalidated data likely contains some invalid samples, they will need to be filtered further as more clips are validated.
We also provide the unfiltered train data so that corpus users can compile their own filter rules.

\subsection{Corpus Structure}

We provide randomly generated train, validation, and test splits, ensuring that each speaker is part of only one split.
The target size of the validation and test splits is 5.3 hours each.
\autoref{tab:splits} shows the number of hours, sentences, and speakers of each split.
To ensure optimal quality, validation and test splits only contain validated samples.
Furthermore, to obtain balanced sets and a larger variety of speakers, we only allow speakers with 5 to 200 recorded sentences to be part of either validation or test splits.

\subsection{Data Statistics}

\begin{figure}[!t]
    \centering
    \includegraphics[width=0.49\textwidth]{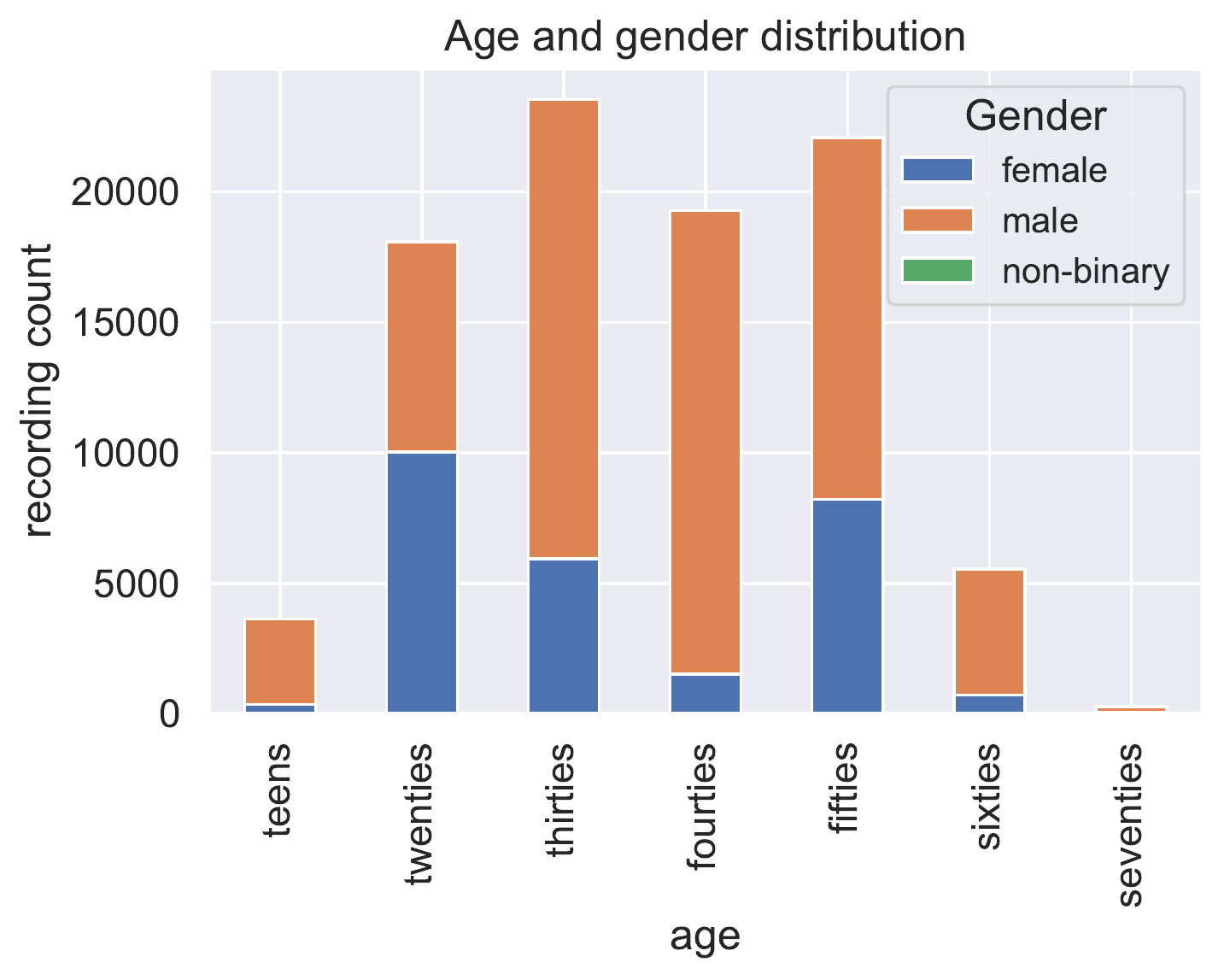}
    \caption{Number of utterances per speaker's age group and gender.}
    \label{fig:age_dist}
\end{figure}

\begin{figure}[!t]
    \centering
    \includegraphics[width=0.49\textwidth]{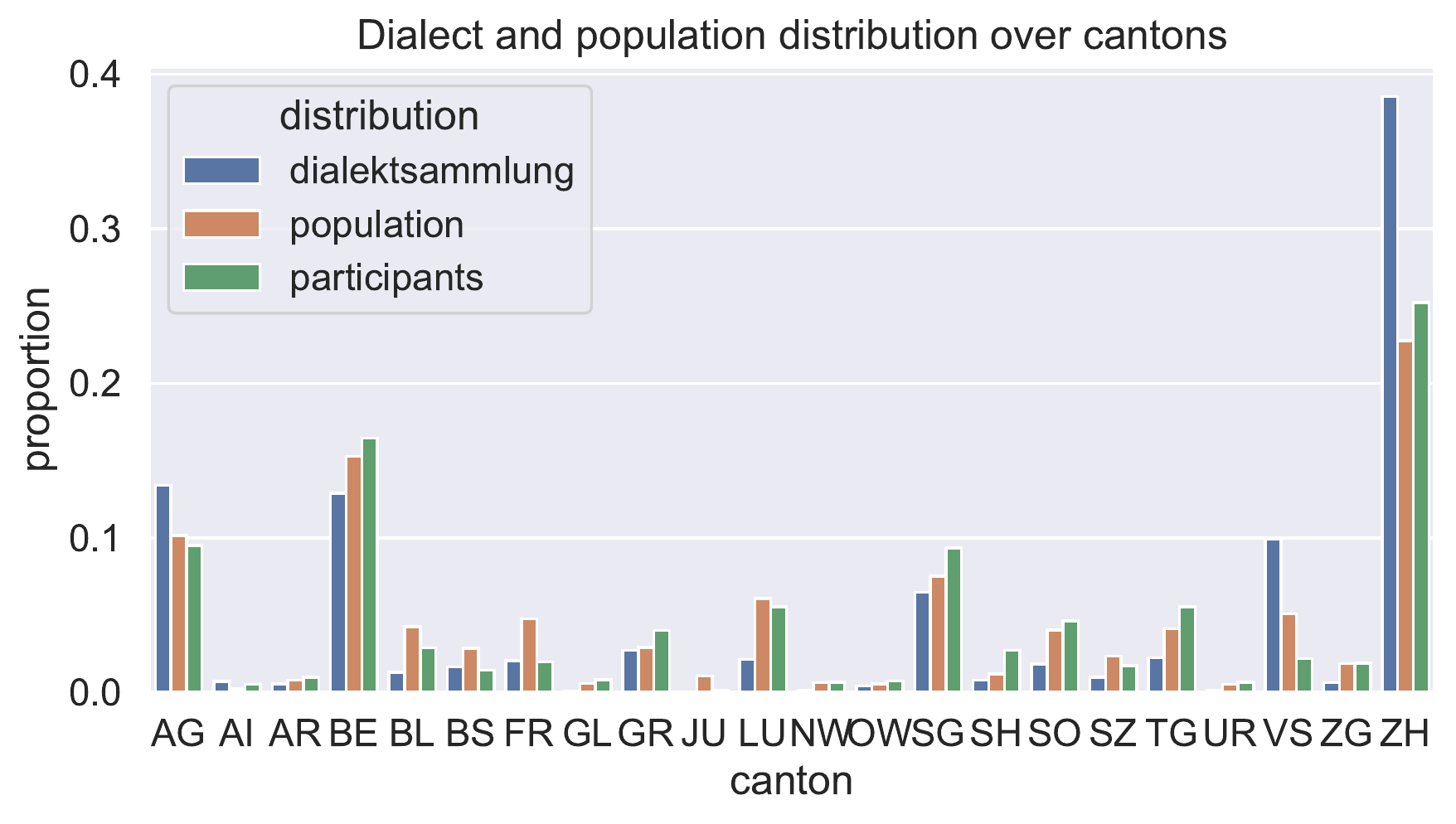}
    \caption{Canton distribution in the dataset compared with the relative population and the relative number of unique speakers per respective canton. Only cantons where Swiss German is spoken are shown.}
    \label{fig:dialect_dist}
\end{figure}

\begin{figure}[!t]
    \centering
    \includegraphics[width=0.49\textwidth]{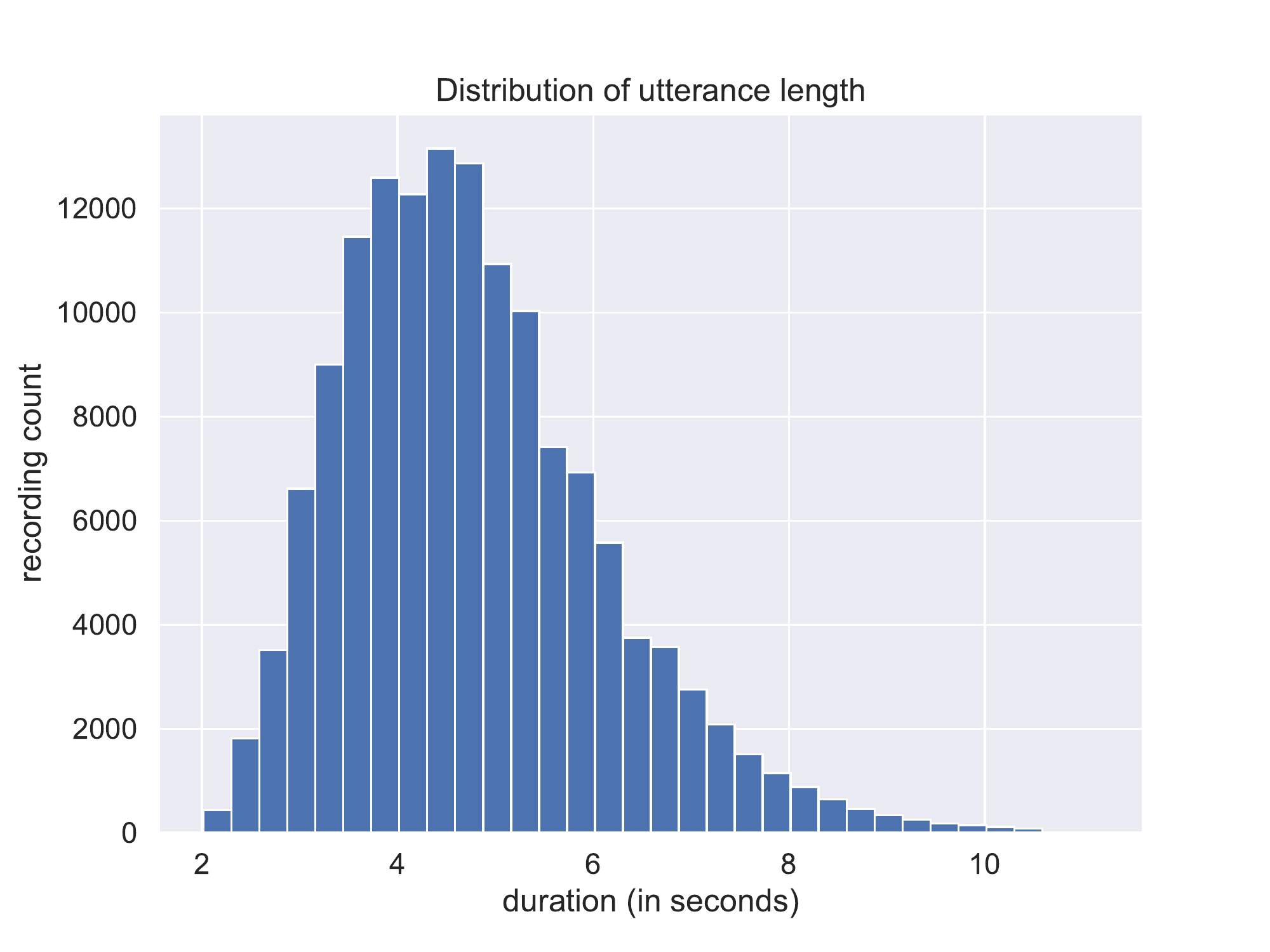}
    \caption{Distribution of utterance lengths in the \CORPUSNAME.}
    \label{fig:utterance_dist}
\end{figure}

On average, an utterance is 4.8 seconds long with a standard deviation of 1.3 seconds. The shortest and longest utterances are 2 and 11.2 seconds long, respectively.
In Figure \ref{fig:utterance_dist} we display the utterance length distribution.

\begin{table*}[!t]
    \centering
    \begin{tabular}{lcccccc}
        \toprule 
        \multirow{2}{*}{\textbf{Model}} & \multirow{2}{*}{\textbf{Train data}} & \multirow{2}{*}{\shortstack{\textbf{Model}\\\textbf{parameters}}} & \multicolumn{2}{c}{WER} & \multicolumn{2}{c}{BLEU} \\
         & & & \textbf{valid} & \textbf{test} & \textbf{valid} & \textbf{test} \\ 
         \midrule
        Transformer & \CORPUSNAME & $72$M & $31.3$ & $30.3$ & $52.1$ & $53.1$ \\
        Transformer & \CORPUSNAME+SPC & $72$M & $24.9$ & $24.7$ & $60.9$ & $61.0$ \\
         \midrule
        XLS-R (0.3B) & \CORPUSNAME & $317$M & $27.2$ & $26.9$ & $54.9$ & $54.6$ \\
        XLS-R (1B) & \CORPUSNAME & $965$M & $21.7$ & $21.6$ & $63.9$ & $64.0$\\
        \bottomrule
    \end{tabular}
    \caption{Performance of the Transformer Baseline and XLS-R Wav2Vec models finetuned on the \CORPUSNAME~train set. We report Word Error Rate (WER) and BLEU scores obtained from evaluating on the \CORPUSNAME~valid and test splits.}
    \label{tab:results_full}
\end{table*}

By crowdsourcing the data, we obtain a diverse set of speakers regarding age, gender, and dialect.
In total, the filtered \CORPUSNAME~contains 142'545 utterances with 138'553 unique sentences.
The vocabulary consists of 41'289 German words.
Out of 3816 speakers, 8\% are male, 6\% are female, 86\% did not reveal their gender, and 4 participants are non-binary.
In terms of utterances, 19\% of utterances are voiced by females, 46\% by males, and 35\% of unknown gender.
On average, each participant recorded 37 utterances with a standard deviation of 364 utterances.
The participant with the most speech donations recorded 13'333 utterances.
In Figure \ref{fig:age_dist} we display the age and gender distribution over the recorded utterances.
In Figure \ref{fig:dialect_dist} we show the distribution over the number of recordings for each canton and compare them with the population of the respective cantons\footnote{We use the canton information as an indicator for the dialect.} and the proportion of unique speakers.
The collected dialects follow the dialect distribution in Switzerland closely, with some exceptions.
For Appenzell Innerrhoden, we have four times more utterances than the relative population. Wallis and Zürich have almost twice as many utterances.
In the canton Wallis, one speaker recorded 10'368 out of 11'739 samples.
The cantons Baselland, Glarus, Jura, Luzern, Nidwalden, Uri, and Zug are underrepresented in the \CORPUSNAME.

\section{Baseline}

We conducted experiments to demonstrate the use of the \CORPUSNAME~corpus for speech translation. We further evaluated how the corpus can be combined with the SPC~\cite{pluess2021a}. Finally, we assessed how large-scale pre-training on unlabeled speech data can improve the performance by finetuning XLS-R Wav2vec models~\cite{babu2021xlsr} on the \CORPUSNAME~train set. 

\noindent{\textbf{Transformer Baseline.}} We employed Transformer~\cite{vaswani2017attention} based models implemented in the FAIRSEQ S2T library~\cite{ott2019fairseq,wang2020fairseqs2t} as our baselines. These models consist of a two-layer convolutional subsampler followed by a Transformer network with 12 encoder layers and six decoder layers. For the Transformer network, we employed eight attention heads, an embedding dimension size of 512, and a dropout rate of 0.15. We used the default model hyper-parameters and learning rate schedules provided by the library without any task-specific tuning.
We evaluated the model performance when training on \CORPUSNAME~ alone as well as the combination of \CORPUSNAME~ and the SPC. After training, we averaged the weights of the ten checkpoints with the lowest validation loss to obtain the final model.

\noindent{\textbf{XLS-R fine-tuning.}} For the Wav2vec experiments, we employed XLS-R models~\cite{babu2021xlsr} that were pre-trained on 436K hours of unlabeled speech data covering more than 128 languages and are publicly available\footnote{\url{https://github.com/pytorch/fairseq/tree/main/examples/wav2vec/xlsr}}.
Importantly, Swiss German was not part of the training data. Of the available pre-trained models, we evaluated XLS-R (0.3B) and XLS-R (1B), whereas the number in braces denotes the number of model parameters. XLS-R Wav2vec models consist of a convolutional feature encoder, followed by a stack of transformer blocks. Details of the architecture configurations can be found in~\cite{babu2021xlsr}. For the finetuning on the \CORPUSNAME~corpus, we followed the procedure and hyper-parameters described by the authors. 

\noindent{\textbf{Results.}} The results of our experiments are shown in \autoref{tab:results_full}.  Both additional labeled training data and large-scale self-supervised pre-training on unlabeled speech data lead to performance improvements. The strong performance of XLS-R (0.3B) highlights the benefits of latter in low-resource settings, even if the target language was not available during pre-training.
Notably, for all our experiments, we did not use any external language model.

\section{Conclusion}
In this work, we presented~\CORPUSNAME, a speech translation dataset for Swiss German speech to Standard German text. The main characteristics of this corpus are the large variety of Swiss German dialects that are covered and the large number of speakers that contributed to the data collection. The baseline achieved 30.3 WER score, and 53.1 BLEU score on the \CORPUSNAME~test set. The current version contains around 200 hours of speech. 

Our goal is to increase the size of the corpus in the future, which will allow for even better performance. We plan to find new ways to engage the public, for instance, by adding gamification components to keep the engagement high. The current version is publicly available. 

\section{Acknowledgements}
First and foremost, we would like to thank all participants for their contribution to this corpus. Furthermore, we thank Tamedia for providing the newspaper texts. We also thank Claudio Zuccolini, Frölein da Capo, Mike Müller, Renato Kaiser, Simon Enzler, and Sina for their public relations efforts.

\section{Bibliographical References}\label{reference}

\bibliographystyle{lrec2022-bib}
\bibliography{main}

\begin{thebibliography}{}

\bibitem[\protect\citename{Ardila \bgroup et al.\egroup }2020]{ardila2020}
Ardila, R., Branson, M., Davis, K., Henretty, M., Kohler, M., Meyer, J.,
  Morais, R., Saunders, L., Tyers, F.~M., and Weber, G.
\newblock (2020).
\newblock {Common Voice: A Massively-Multilingual Speech Corpus}.
\newblock In {\em Proceedings of the 12th Conference on Language Resources and
  Evaluation (LREC 2020)}, pages 4211--4215.

\bibitem[\protect\citename{Babu \bgroup et al.\egroup }2021]{babu2021xlsr}
Babu, A., Wang, C., Tjandra, A., Lakhotia, K., Xu, Q., Goyal, N., Singh, K.,
  von Platen, P., Saraf, Y., Pino, J., Baevski, A., Conneau, A., and Auli, M.
\newblock (2021).
\newblock Xls-r: Self-supervised cross-lingual speech representation learning
  at scale.
\newblock {\em arXiv}, abs/2111.09296.

\bibitem[\protect\citename{Di~Gangi \bgroup et al.\egroup
  }2019]{di-gangi-etal-2019-must}
Di~Gangi, M.~A., Cattoni, R., Bentivogli, L., Negri, M., and Turchi, M.
\newblock (2019).
\newblock {M}u{ST}-{C}: a {M}ultilingual {S}peech {T}ranslation {C}orpus.
\newblock In {\em Proceedings of the 2019 Conference of the North {A}merican
  Chapter of the Association for Computational Linguistics: Human Language
  Technologies, Volume 1 (Long and Short Papers)}, pages 2012--2017,
  Minneapolis, Minnesota, June. Association for Computational Linguistics.

\bibitem[\protect\citename{Dogan-Schönberger \bgroup et al.\egroup
  }2021]{Dogan2021}
Dogan-Schönberger, P., Mäder, J., and Hofmann, T.
\newblock (2021).
\newblock {SwissDial: Parallel Multidialectal Corpus of Spoken Swiss German}.

\bibitem[\protect\citename{Garner \bgroup et al.\egroup }2014]{Garner2014b}
Garner, P.~N., Imseng, D., and Meyer, T.
\newblock (2014).
\newblock {Automatic Speech Recognition and Translation of a {S}wiss {G}erman
  Dialect: {W}alliserdeutsch}.
\newblock In {\em Proceedings of Interspeech}, Singapore, September.

\bibitem[\protect\citename{Iranzo-Sánchez \bgroup et al.\egroup
  }2020]{9054626}
Iranzo-Sánchez, J., Silvestre-Cerdà, J.~A., Jorge, J., Roselló, N.,
  Giménez, A., Sanchis, A., Civera, J., and Juan, A.
\newblock (2020).
\newblock {Europarl-ST: A Multilingual Corpus for Speech Translation of
  Parliamentary Debates}.
\newblock In {\em ICASSP 2020 - 2020 IEEE International Conference on
  Acoustics, Speech and Signal Processing (ICASSP)}, pages 8229--8233.

\bibitem[\protect\citename{Needleman and Wunsch}1970]{Needleman-Wunsch}
Needleman, S.~B. and Wunsch, C.~D.
\newblock (1970).
\newblock {A general method applicable to the search for similarities in the
  amino acid sequence of two proteins}.
\newblock {\em J. Mol. Biol.}, 48:443--453.

\bibitem[\protect\citename{Ott \bgroup et al.\egroup }2019]{ott2019fairseq}
Ott, M., Edunov, S., Baevski, A., Fan, A., Gross, S., Ng, N., Grangier, D., and
  Auli, M.
\newblock (2019).
\newblock {fairseq: A Fast, Extensible Toolkit for Sequence Modeling}.
\newblock In {\em Proceedings of NAACL-HLT 2019: Demonstrations}.

\bibitem[\protect\citename{Pl{\"u}ss \bgroup et al.\egroup }2021]{pluess2021a}
Pl{\"u}ss, M., Neukom, L., Scheller, C., and Vogel, M.
\newblock (2021).
\newblock {Swiss Parliaments Corpus, an Automatically Aligned Swiss German
  Speech to Standard German Text Corpus}.
\newblock In {\em Swiss Text Analytics Conference 2021}, Proceedings of the
  Swiss Text Analytics Conference 2021.

\bibitem[\protect\citename{Scherrer \bgroup et al.\egroup }2019]{Scherrer2019}
Scherrer, Y., Samard{\v{z}}i{\'{c}}, T., and Glaser, E.
\newblock (2019).
\newblock {ArchiMob: Ein multidialektales Korpus schweizerdeutscher
  Spontansprache}.
\newblock {\em Linguistik Online}, 98(5):425--454, November.

\bibitem[\protect\citename{Vaswani \bgroup et al.\egroup
  }2017]{vaswani2017attention}
Vaswani, A., Shazeer, N., Parmar, N., Uszkoreit, J., Jones, L., Gomez, A.~N.,
  Kaiser, L.~u., and Polosukhin, I.
\newblock (2017).
\newblock {Attention is All you Need}.
\newblock In {\em Advances in Neural Information Processing Systems},
  volume~30. Curran Associates, Inc.

\bibitem[\protect\citename{Wang \bgroup et al.\egroup
  }2020]{wang2020fairseqs2t}
Wang, C., Tang, Y., Ma, X., Wu, A., Okhonko, D., and Pino, J.
\newblock (2020).
\newblock {fairseq S2T: Fast Speech-to-Text Modeling with fairseq}.
\newblock In {\em Proceedings of the 2020 Conference of the Asian Chapter of
  the Association for Computational Linguistics (AACL): System Demonstrations}.

\end{thebibliography}


\end{document}